\documentclass{article}
\usepackage{microtype}
\usepackage{graphicx}
\usepackage{booktabs} 
\usepackage{multirow}
\usepackage{multicol}
\usepackage{tablefootnote}
\usepackage{xcolor}
\usepackage{fix-cm} 
\DeclareMathSizes{10}{10}{6}{4}
\usepackage{hyperref}

\usepackage{amsmath}
\usepackage{amssymb}
\usepackage{mathtools}
\usepackage{amsthm}
\usepackage{caption}
\usepackage{subcaption}
\usepackage{booktabs}
\usepackage[flushleft]{threeparttable}
\usepackage{MnSymbol}
\usepackage{lipsum}

\usepackage{enumitem}
\usepackage[authoryear]{natbib}
\usepackage{algorithm}
\usepackage{algpseudocode}

\algrenewcommand\algorithmicrequire{\textbf{Input:}}
\algrenewcommand\algorithmicensure{\textbf{Output:}}
\algrenewcommand\algorithmiccomment[1]{\hfill\textcolor{blue}{$\triangleright$\ #1}}

\usepackage{mathtools}
\usepackage{array}
\usepackage{lipsum}
\usepackage{url}
\usepackage{tablefootnote}
\usepackage[export]{adjustbox}
\usepackage{footmisc}
\usepackage{balance}
\usepackage{amsmath,amssymb,amsfonts}%
\usepackage{bm}

\usepackage{contour}
\usepackage[bb=dsserif]{mathalpha}
\usepackage{bm}
\usepackage{bigstrut}
\usepackage{dirtytalk}
\usepackage{enumitem}
\usepackage{colortbl}
\usepackage{xurl}
\usepackage{soul}
\usepackage{adjustbox}
\usepackage{tcolorbox}
\usepackage{ulem}

\usepackage[capitalize,noabbrev]{cleveref}
\crefformat{footnote}{#2\footnotemark[#1]#3}

\makeatletter
\newtheorem*{rep@theorem}{\rep@title}
\newcommand{\newreptheorem}[2]{%
\newenvironment{rep#1}[1]{%
 \def\rep@title{#2 \ref{##1}}%
 \begin{rep@theorem}}%
 {\end{rep@theorem}}}
\makeatother

\theoremstyle{plain}

\newreptheorem{theorem}{Theorem}

\newreptheorem{lemma}{Lemma}

\newreptheorem{corollary}{corollary}
\theoremstyle{definition}

\theoremstyle{remark}

\usepackage[textsize=tiny]{todonotes}

\def\mydefbb#1{\expandafter\def\csname bb#1\endcsname{\ensuremath{\mathbb{#1}}}}
\def\mydefallbb#1{\ifx#1\mydefallbb\else\mydefbb#1\expandafter\mydefallbb\fi}
\mydefallbb ABCDEFGHIJKLMNOPQRSTUVWXYZ\mydefallbb

\def\mydefcal#1{\expandafter\def\csname cal#1\endcsname{\ensuremath{\mathcal{#1}}}}
\def\mydefallcal#1{\ifx#1\mydefallcal\else\mydefcal#1\expandafter\mydefallcal\fi}
\mydefallcal ABCDEFGHIJKLMNOPQRSTUVWXYZ\mydefallcal

\makeatletter
\let\save@mathaccent\mathaccent
\newcommand*\if@single[3]{%
  \setbox0\hbox{${\mathaccent"0362{#1}}^H$}%
  \setbox2\hbox{${\mathaccent"0362{\kern0pt#1}}^H$}%
  \ifdim\ht0=\ht2 #3\else #2\fi
  }
\newcommand*\rel@kern[1]{\kern#1\dimexpr\macc@kerna}
\newcommand*\widebar[1]{\@ifnextchar^{{\wide@bar{#1}{0}}}{\wide@bar{#1}{1}}}
\newcommand*\wide@bar[2]{\if@single{#1}{\wide@bar@{#1}{#2}{1}}{\wide@bar@{#1}{#2}{2}}}
\newcommand*\wide@bar@[3]{%
  \begingroup
  \def\mathaccent##1##2{%
    \let\mathaccent\save@mathaccent
    \if#32 \let\macc@nucleus\first@char \fi
    \setbox\z@\hbox{$\macc@style{\macc@nucleus}_{}$}%
    \setbox\tw@\hbox{$\macc@style{\macc@nucleus}{}_{}$}%
    \dimen@\wd\tw@
    \advance\dimen@-\wd\z@
    \divide\dimen@ 3
    \@tempdima\wd\tw@
    \advance\@tempdima-\scriptspace
    \divide\@tempdima 10
    \advance\dimen@-\@tempdima
    \ifdim\dimen@>\z@ \dimen@0pt\fi
    \rel@kern{0.6}\kern-\dimen@
    \if#31
      \overline{\rel@kern{-0.6}\kern\dimen@\macc@nucleus\rel@kern{0.4}\kern\dimen@}%
      \advance\dimen@0.4\dimexpr\macc@kerna
      \let\final@kern#2%
      \ifdim\dimen@<\z@ \let\final@kern1\fi
      \if\final@kern1 \kern-\dimen@\fi
    \else
      \overline{\rel@kern{-0.6}\kern\dimen@#1}%
    \fi
  }%
  \macc@depth\@ne
  \let\math@bgroup\@empty \let\math@egroup\macc@set@skewchar
  \mathsurround\z@ \frozen@everymath{\mathgroup\macc@group\relax}%
  \macc@set@skewchar\relax
  \let\mathaccentV\macc@nested@a
  \if#31
    \macc@nested@a\relax111{#1}%
  \else
    \def\gobble@till@marker##1\endmarker{}%
    \futurelet\first@char\gobble@till@marker#1\endmarker
    \ifcat\noexpand\first@char A\else
      \def\first@char{}%
    \fi
    \macc@nested@a\relax111{\first@char}%
  \fi
  \endgroup
}
\makeatother

\definecolor{kjred}{RGB}{228, 26, 28}
\definecolor{kjblue}{RGB}{55,126,184}
\definecolor{kjgreen}{RGB}{77,175,74}
\definecolor{kjpurple}{RGB}{152,78,163}
\definecolor{kjorange}{RGB}{255,127,0}

\definecolor{Red}{RGB}{244, 124, 124}
\definecolor{Green}{RGB}{162, 222, 147}
\definecolor{Blue}{RGB}{112, 161, 211}

\newcommand{\smallsection}[1]{{\noindent {\bf{\underline{\smash{#1}}}}}}

\definecolor{peace}{RGB}{228, 26, 28}
\definecolor{love}{RGB}{55, 126, 184}
\definecolor{joy}{RGB}{77, 175, 74}
\definecolor{kindness}{RGB}{152, 78, 163}



\newcommand{\citenameyear}[1]{\citeauthor{#1} \citeyearpar{#1}}

\renewcommand{\cite}{\citenameyear}

\title{On Training-Test (Mis)alignment in Unsupervised Combinatorial Optimization: Observation, Empirical Exploration, and Analysis}

\author{
Fanchen Bu\footnote{School of Electrical Engineering, KAIST, Daejeon, Republic of Korea; AI4CO Open-Source Community; boqvezen97@kaist.ac.kr} \quad
Kijung Shin\footnote{Kim Jaechul Graduate School of Artificial Intelligence, KAIST, Seoul, Republic of Korea; School of Electrical Engineering, KAIST, Daejeon, Republic of Korea; kijungs@kaist.ac.kr}
}

\date{}

\begin{document}

\maketitle

\begin{abstract}
In \textit{unsupervised combinatorial optimization} (UCO), during training, one aims to have continuous decisions that are promising in a \textit{probabilistic} sense for each training instance, which enables end-to-end training on initially discrete and non-differentiable problems.
At the test time, for each test instance, starting from continuous decisions, \textit{derandomization} is typically applied to obtain the final deterministic decisions.
Researchers have developed more and more powerful test-time derandomization schemes to enhance the empirical performance and the theoretical guarantee of UCO methods.
However, we notice a misalignment between training and testing in the existing UCO methods.
Consequently, lower training losses do not necessarily entail better post-derandomization performance, \textit{even for the training instances without any data distribution shift}.
Empirically, we indeed observe such undesirable cases.
We explore a preliminary idea to better align training and testing in UCO by including a differentiable version of derandomization into training.
Our empirical exploration shows that such an idea indeed improves training-test alignment, but also introduces nontrivial challenges into training.

\end{abstract}

\section{Introduction}\label{sec:intro}

Combinatorial optimization (CO) problems are naturally discrete.
Typical examples include optimization problems on graphs where we make binary yes-or-no decisions on each node, and the objective is a function of the graph structure and the binary decisions.
CO problems have a rich lineage in various research fields, including theoretical computer science~\citep{arumugam2016handbook} and operations research~\citep{modaresi2020learning}, with real-world applications from network design~\citep{cheng2006combinatorial} to scheduling~\citep{hwang2001combinatorial} and bioinformatics~\citep{bauer2007accurate}.
However, the discrete nature of CO problems makes it non-trivial to apply typical machine learning methods that are based on differentiable optimization to them.

To overcome these challenges, researchers have explored various strategies to effectively combine machine learning with CO problems~\citep{bengio2021machine}. 
Early approaches often involved supervised methods~\citep{li2018combinatorial} or reinforcement learning techniques~\citep{kool2018attention,berto2023rl4co}. 
However, these methods require labeled data or extensive interaction, limiting their applicability and generalizability in many real-world scenarios~\citep{karalias2020erdos}. Consequently, the field has seen a growing interest in unsupervised approaches, resulting in the development of \textit{unsupervised combinatorial optimization} (UCO) methods.

The key idea of UCO is to (1) allow continuous decisions and (2) evaluate the expected objective by interpreting the continuous decisions as random variables, which gives a fully differentiable process and enables end-to-end training.
At the test time, \textit{derandomization} is typically applied to continuous decisions to transform them into deterministic decisions as the final output~\citep{karalias2020erdos}.

Over the course of time, UCO researchers have developed more and more powerful test-time derandomization schemes, from naive random sampling~\citep{karalias2020erdos}, iterative rounding~\citep{karalias2020erdos,wang2022unsupervised}, to greedy derandomization~\citep{Bu2024UCom2}.
With the development of test-time derandomization, we have witnessed the enhancement of the empirical performance as well as the theoretical quality guarantee of UCO methods.

However, we notice a misalignment between training and testing in the existing UCO methods: the training essentially tries to optimize the expected quality of the output continuous decisions assuming naive random sampling, while rather sophisticated derandomization is actually used at the test time.
Therefore, even if we have lower training losses, we cannot guarantee better post-derandomization performance at the test time, \textit{even for the training instances} (i.e., here we are not discussing the training-test misalignment regarding data distributions but regarding methodology).
We indeed empirically observe such undesirable cases.

We explore a preliminary idea to better align training and testing in UCO by including a differentiable version of derandomization into training.
By our empirical exploration, we validate that such an idea indeed can improve training-test alignment.
However, we also observe that including such additional soft derandomization schemes into training increases the difficulty of training, i.e., we may not be able to have the training losses stably decrease during training.
Our analysis suggests that the future development of UCO methods may need to find a balance between training-test alignment and the ease of training.

\smallsection{Reproducibility.}
The code and datasets are available in the online appendix~\citep{appendix}.\footnote{\url{https://github.com/bokveizen/uco_derand}}
\begin{algorithm}[t]
\caption{Iterative Rounding}
\label{alg:iterative-rounding}
\begin{algorithmic}[1]
  \Require     
    \begin{tabular}[t]{@{}r@{}l@{}}
             (1) & ~Continuous decisions $\tilde{D}\in[0,1]^n$,\\
             (2) & ~Rounding sequence    $\pi_n=(v_1,\dots,v_n)$
           \end{tabular}
  \Ensure Final discrete decisions $D \in\{0,1\}^n$
  \State $D \gets \tilde{D}$ \Comment{Initialization}
  \For{$j \gets v_1 \,\mathbf{to}\, v_n$} \Comment{Iteration following the sequence}
    \For{$b\in\{0,1\}$}
      \State $D' \gets D$ \Comment{Copy}
      \State $D'_{j} \gets b$ \Comment{Modify a single entry}
      \State $\Delta_{b} \gets \tilde f(D) - \tilde f(D')$ \Comment{Evaluation}
    \EndFor
    \State $D_{j} \gets \arg\max_{b\in\{0,1\}} \Delta_b$ \Comment{Rounding}
  \EndFor
  \State \Return $D$
\end{algorithmic}
\end{algorithm}

\section{Preliminaries and Background}\label{sec:prelim}

\subsection{Combinatorial Optimization (CO)}\label{subsec:CO}

We consider CO problems with $n$ binary decisions, where $n$ is a positive integer.
A CO problem aims to find the \textit{optimal decisions} $D_{\textrm{opt}}$ inside a \textit{feasible set} $\calC$ such that $D_{\textrm{opt}} \in \calC \subseteq \{0,1\}^n$, to minimize an \textit{objective} $f: \calC \mapsto \mathbb{R}$ (i.e., $D_{\textrm{opt}} = \arg \min_{D \in \calC} f(D)$).

\subsection{Unsupervised Combinatorial Optimization (UCO)}\label{subsec:UCO}

Based on the probabilistic method~\citep{erdos1974probabilistic}, \cite{karalias2020erdos} proposed the UCO framework with the high-level idea to 
(1) apply continuous relaxation to the objective $f$ and its domain, to obtain $\tilde{f}: [0,1]^n \to \mathbb{R}$ and enable end-to-end training, and 
(2) apply derandomization at the test time to obtain the final output decisions $D_{\mathrm{out}} \in \calC$.

\smallsection{Continuous relaxation.} Each $\tilde{D} \in [0,1]^n$ is interpreted as a \textit{distribution} (typically, an independent multivariate Bernoulli distribution) on $\{0,1\}^n$,
and we aim to construct a \textit{differentiable} $\tilde{f}$ such that $\tilde{f}(\tilde{D}) \approx \mathbb{E}_{D\sim\tilde{D}}[f(D)] + \beta \Pr_{D \sim \tilde{D}}[D \notin \calC]$ with constraint coefficient $\beta > 0$.
The key points are 
(1) now we are able to conduct end-to-end training with this differentiable surrogate $\tilde{f}$ instead of the originally discrete non-differentiable $f$, and 
(2) when $\tilde{f}$ is minimized, it is guaranteed that the optimal $\tilde{D}_{\mathrm{opt}} = \arg \min_{\tilde{D} \in [0,1]^n} \tilde{f}(\tilde{D})$ corresponds to the optimal ${D}_{\mathrm{opt}}$ for the original objective $f$.

\begin{algorithm}[t]
\caption{Greedy Rounding}
\label{alg:greedy-rounding}
\begin{algorithmic}[1]
  \Require Continuous decisions $\tilde{D}\in[0,1]^n$
  \Ensure Final discrete decisions $D\in\{0,1\}^n$
  \State $D \gets \tilde{D}$ \Comment{Initialization}
  \Repeat
    \For{$j\in\{1,\dots,n\}$ and $b\in\{0,1\}$}
      \State $D' \gets D$ \Comment{Copy}
      \State $D'_j \gets b$ \Comment{Modify a single entry}
      \State $\Delta_{j,b} \gets \tilde f(D) - \tilde f(D')$ \Comment{Evaluation}
    \EndFor
    \State $(j^*, b^*) \gets \arg \max_{j, b \in [n] \times \{0, 1\}} \Delta_{j,b}$ \Comment{Best choice}
    \State $D_{j^*} \gets b^*$ \Comment{Rounding with the best choice}    
  \Until{$\Delta_{j^*, b^*}\leq 0$} \Comment{Until local minimum}
  \State \Return $D$
\end{algorithmic}
\end{algorithm}

\smallsection{Derandomization.}
At the test time, for each test instance, \textit{derandomization} is used to obtain the final output discrete decisions.
Researchers have considered various derandomization schemes for UCO, including (let $\tilde{D}_{\mathrm{output}} \in [0,1]^n$ be the initial continuous output for the test instance):
\begin{itemize}[leftmargin=*,topsep=0pt,itemsep=0pt]
    \item \textbf{Naive random sampling} \citep{karalias2020erdos}: We sample $D_{\mathrm{out}}$ from the distribution represented by $\tilde{D}_{\mathrm{output}}$;
    \item \textbf{Iterative rounding} \citep{karalias2020erdos,wang2022unsupervised}: We fix a sequence $\pi_n = \{v_1, v_2, \ldots, v_n\}$, iterate for $i = 1,2,\ldots, n$ while rounding for $v_i$ to the one between $0$ and $1$ that gives lower $\tilde{f}$, and obtain $D_{\mathrm{out}}$ after rounding all the $n$ entries (see Algorithm~\ref{alg:iterative-rounding});
    \item \textbf{Greedy rounding} \citep{Bu2024UCom2}: Repeatedly, we consider all the $n \times 2$ possible rounding possibilities (first choose an entry $i \in [n]$ and then decide to round it to $0$ or $1$) and greedily pick the one that gives the lowest value for $\tilde{f}$, until no further rounding can improve $\tilde{f}$, i.e., when reaching a local minimum (see Algorithm~\ref{alg:greedy-rounding}).
\end{itemize}
To conclude, over the course of time, more and more powerful derandomization schemes have been proposed, and we have observed that such schemes improve both theoretical guarantees and empirical performance in UCO.
\begin{table*}[t]
\centering
\resizebox{\textwidth}{!}{%
\begin{tabular}{lcccccc}
\toprule
Trial     & 1 & 2 & 3 & 4 & 5 & Average \\
\midrule
Iterative & 1968 (39.8\%) & 1826 (36.9\%) & 1957 (39.5\%) & 2094 (42.3\%) & 1549 (31.3\%) & 1878.8 (38.0\%) \\
Greedy    & 1958 (39.6\%) & 2252 (45.5\%) & 2299 (46.4\%) & 1974 (39.9\%) & 1321 (26.7\%) & 1960.8 (39.6\%) \\
\bottomrule
\end{tabular}%
}
\caption{\textbf{Empirically, there are many ``bad'' pairs where the surrogate objective and the final post-derandomization objective give different relative ordering.} For each derandomization scheme (iterative or greedy rounding), we report the number of ``bad'' pairs among all the 4950 pairs in each random trial, as well as the average number over all trials.}
\label{tab:bad_cases_toy}
\end{table*}

\section{Observation: Training-Test Misalignment}\label{sec:obs}

Although more powerful derandomization schemes enable better theoretical guarantees and empirical performance, we identify a training-test misalignment issue in existing UCO.
In this section, we shall discuss the issue from a methodological perspective and provide empirical evidence.

\subsection{Methodological Misalignment}\label{sec:obs:method_misalign}
In training, the surrogate objective $\Tilde{f} \approx \mathbb{E}_{D\sim\tilde{D}}[f(D)] + \beta \Pr_{D \sim \tilde{D}}[D \notin \calC]$ essentially evaluates the expected objective (plus penalty on the probability of violating the constraints) when we obtain $D$ by naive random sampling from $\tilde{D}$.
However, at the testing time, existing UCO methods have actually used much more sophisticated derandomization schemes (iterative or greedy rounding).

Although the construction of $\Tilde{f}$ guarantees that the optimal $\tilde{D}_{\mathrm{opt}}$ for $\Tilde{f}$ corresponds to the optimal ${D}_{\mathrm{opt}}$ for the original objective $f$, in principle we cannot guarantee the training actually finds the optimum due to the complexity of objective in many CO problems, many of which are even NP-hard.

Therefore, in practice, when the surrogate objective $\Tilde{f}$ improves during training, i.e., we obtain new continuous decisions $\tilde{D}_{\mathrm{new}}$ that are better than old ones $\tilde{D}_{\mathrm{old}}$ with $\Tilde{f}(\tilde{D}_{\mathrm{new}}) < \Tilde{f}(\tilde{D}_{\mathrm{old}})$, the corresponding post-derandomization decisions ${D}_{\mathrm{new}}$ and ${D}_{\mathrm{old}}$ may \textit{not} satisfy $f({D}_{\mathrm{new}}) < f({D}_{\mathrm{old}})$.
That is, a better surrogate objective does not necessarily give better final test-time performance, \textit{even for the training instance without any distribution shift}, resulting in an undesirable training-test misalignment.

\subsection{Empirical Evidence: Toy Example}\label{subsec:obs:toy}

Below, we provide empirical evidence for the methodological misalignment discussed above.
We consider a toy example of quadratic functions.
Specifically, we consider the objective $f: \{0,1\}^n \mapsto \mathbb{R}$ in the form of $f(D) = \sum_{i,j = 1}^n \alpha_{ij} d_i d_j$, where $D = (d_1, d_2, \ldots, d_n)$.
We consider the simplistic CO problem with objective $f$ and without any constraints (i.e., $\calC = \{0,1\}^n$).
We can construct the exact expectation $\tilde{f}: [0,1]^n \mapsto \mathbb{R}$ of $f$, which is $\tilde{f}(\tilde{D}) = \sum_{i,j = 1}^n \alpha_{ij} \tilde{d}_i \tilde{d}_j$, where $\tilde{D} = (\tilde{d}_1, \tilde{d}_2, \ldots, \tilde{d}_n)$.

Now, we use $n = 50$, sample random $\alpha_{ij}$'s from i.i.d. normal distributions, and also sample 100 random $\tilde{D}^{(k)}$'s for $k = 1,2,\ldots, 100$, where each $\tilde{D}^{(k)}$ follows an independent multivariate uniform distribution between $0$ and $1$ (i.e., each entry in $\tilde{D}^{(k)}$ follows an i.i.d. uniform distribution $0$ and $1$).
For each $\tilde{D}^{(k)}$, we compute its surrogate objective $\tilde{f}^{(k)} := \tilde{f}(\tilde{D}^{(k)})$, its corresponding outputs $D_{\mathrm{iter}}^{(k)}, D_{\mathrm{grd}}^{(k)} \in \{0,1\}^n$ after iterative rounding and greedy rounding respectively, and the corresponding test-time objective ${f}_{\mathrm{iter}}^{(k)} := f(D_{\mathrm{iter}}^{(k)})$ and ${f}_{\mathrm{grd}}^{(k)} := f(D_{\mathrm{grd}}^{(k)})$.
We have 4950 pairs of $(\tilde{D}^{(k_1)}, \tilde{D}^{(k_2)})$'s in total, and we say a pair $(\tilde{D}^{(k_1)}, \tilde{D}^{(k_2)})$ is ``bad'' if 
$(\tilde{f}^{(k_1)} - \tilde{f}^{(k_2)})({f}^{(k_1)} - {f}^{(k_2)})) < 0$, i.e., if the surrogate objective and the final post-derandomization objective give different relatively ordering.

We repeat the above process in five independent random trials, and report the number of ``bad'' pairs when we use iterative rounding or greedy rounding.
As shown in Table~\ref{tab:bad_cases_toy}, there are many ``bad'' pairs, even for such a simplistic CO problem where we do not have constraints and we can construct the exact expectation as the surrogate objective, which provides empirical evidence to the methodological misalignment discussed in Section~\ref{sec:obs:method_misalign}.




\begin{figure}
    \centering
    \includegraphics[width=0.7\linewidth]{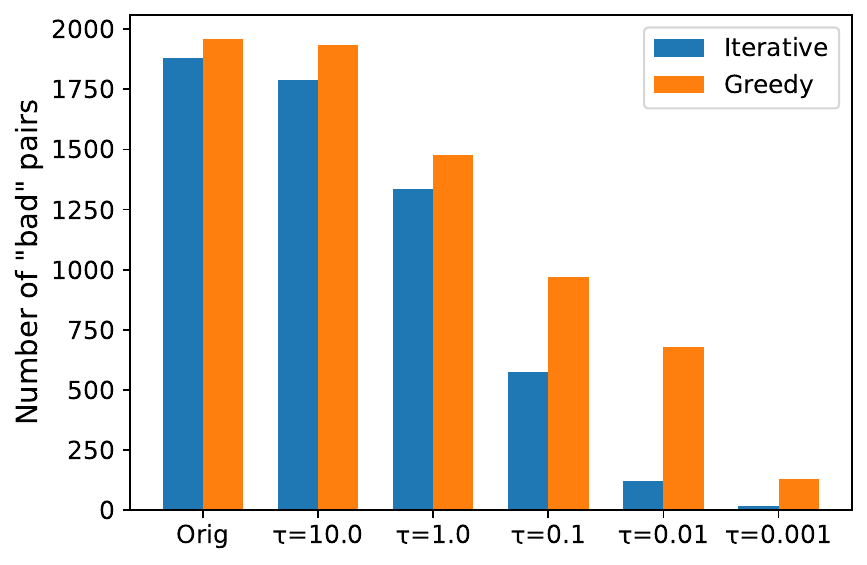}
    \caption{\textbf{Our preliminary idea of including soft derandomization improves training-test alignment.} As the soft temperature $\tau$ decreases, ``bad'' pairs reduces.}
    \label{fig:softmax_toy}
\end{figure}

\begin{figure}[htb]
  \centering
  \begin{subfigure}[t]{0.8\linewidth}
    \centering
    \includegraphics[width=\linewidth]{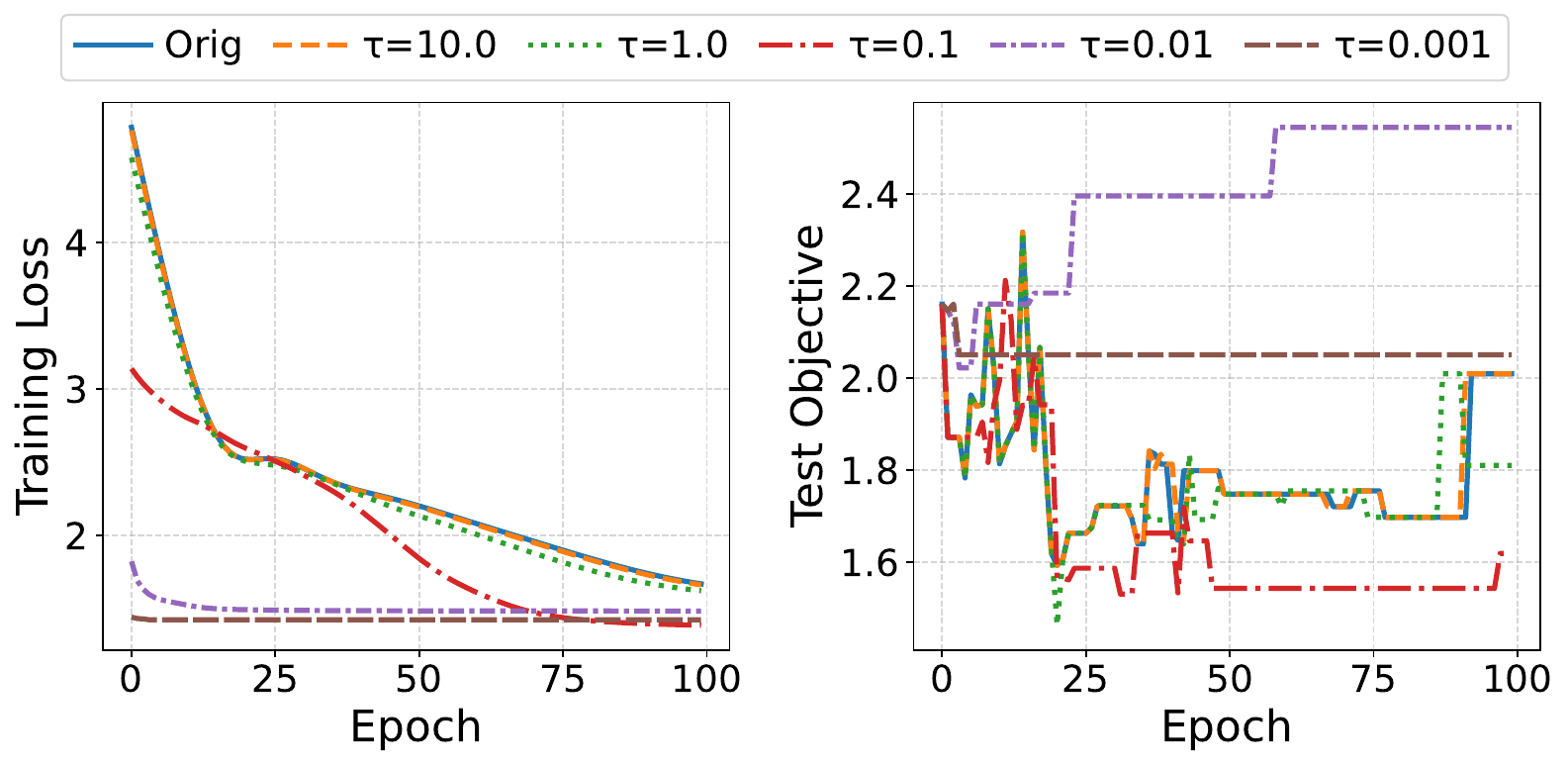}%
    \vspace{-1mm}
    \caption{Iterative rounding}
    \label{fig:fl_curves:iter}
  \end{subfigure}
  \hfill
  \begin{subfigure}[t]{0.8\linewidth
  }
    \centering
    \includegraphics[width=\linewidth]{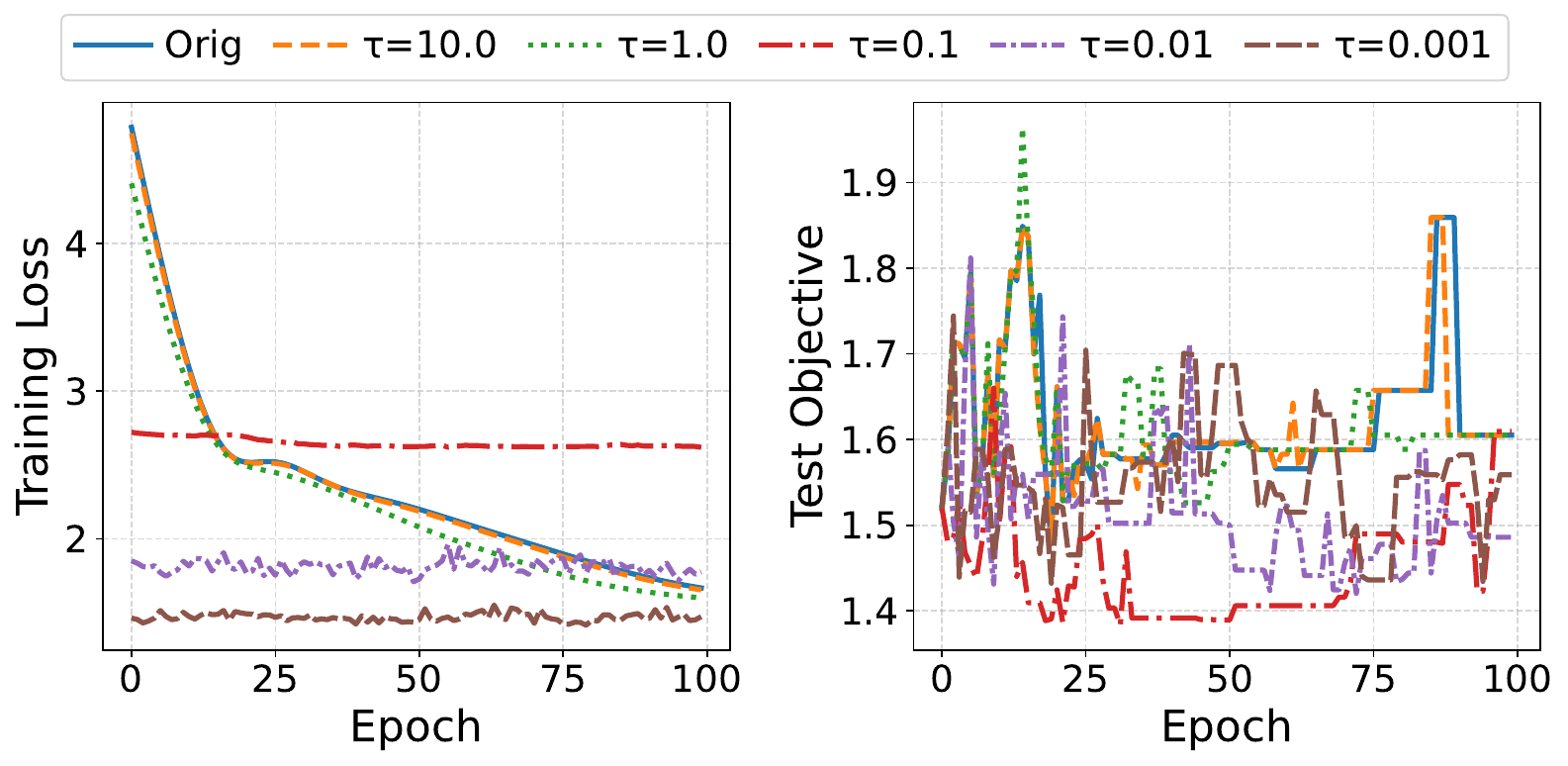}%
    \vspace{-1mm}
    \caption{Greedy rounding}
    \label{fig:fl_curves:greedy}
  \end{subfigure}
  \vspace{-1mm}
  \caption{%
    The curves of training loss and test objective (both lower the better) with soft derandomization using different temperatures $\tau$ on the facility location problem.
  }
  \label{fig:fl_curves}
\end{figure}

\section{Empirical Exploration and Analysis}\label{sec:experiments}

Now that we have identified and empirically observed this training-test misalignment in UCO, how can we improve UCO methods to have better alignment?
Below, we discuss our empirical exploration and analysis, including a preliminary idea to improve training-test alignment in UCO, empirical results, and the challenges we encountered.

\subsection{Preliminary Idea: Soft Derandomization}

Recall the methodological misalignment we identified in Section~\ref{sec:obs:method_misalign}.
The key issue is that the surrogate objective $\Tilde{f} \approx \mathbb{E}_{D\sim\tilde{D}}[f(D)] + \beta \Pr_{D \sim \tilde{D}}[D \notin \calC]$ essentially assumes naive random sampling for training, while more sophisticated derandomization schemes (iterative or greedy rounding) are actually used at the test time.

Therefore, we propose a preliminary and straightforward idea to improve training-test alignment in UCO, which includes (a soft and differentiable version of) the test-time derandomization scheme also in the training.
Specifically, we replace the discrete $\arg \max$ in both iterative rounding (Algorithm~\ref{alg:iterative-rounding}) and greedy rounding (Algorithm~\ref{alg:greedy-rounding}) with a soft differentiable $\operatorname{softmax}$.
It is easy to see that such soft versions are expected to make training and test phases methodologically more similar, and they approach the original derandomization schemes as the temperature decreases.

\subsection{Toy Example}

We first revisit the quadratic-function toy example in Section~\ref{subsec:obs:toy}.
Now, for each continuous $\Tilde{D}$, we first apply the soft version of iterative or greedy rounding to it, before evaluating it with $\Tilde{f}$.
We keep all the other settings the same as in Section~\ref{subsec:obs:toy}, with different softmax temperatures $\tau \in \{10.0, 1.0, 0.1, 0.01, 0.001\}$.

As shown in Figure~\ref{fig:softmax_toy}, we see that the soft version of rounding indeed improves the training-test alignment and reduces the ``bad'' pairs, and lower temperatures give better alignment, validating the correctness of our preliminary idea.

\subsection{Typical CO Problem: Facility Location}

Although our preliminary idea can indeed improve training-test alignment, that is \textit{not} all we need.
Specifically, we need both (1) training-test alignment, i.e., test performance improves as training objective improves, and (2) that training objective actually improves during training.
Therefore, now the question is about the second point: With soft derandomization, does the objective improve during training?

We study a typical CO problem used in UCO literature, facility location~\citep{drezner2004facility}, where we are given a set of locations and we aim to pick a subset of centers to minimize the total distance from each location to its closest picked center.
We follow the experimental settings by~\citep{Bu2024UCom2}, but only check the training and test \textit{on the same instance}, because we are studying training-test (mis)alignment regarding methodology instead of data distributions, as mentioned in Section~\ref{sec:intro}.

In Figure~\ref{fig:fl_curves}, we show the curves of training surrogate objectives and test performance with soft derandomization using different softmax temperatures.
When the temperature $\tau$ is too high (e.g., $\tau = 10.0$), soft derandomization is weak and has negligible effects.
On the other hand, when $\tau$ is too low (e.g., $\tau = 0.01$ or $0.001$), the loss surface has higher curvature and less smooth gradients, and the training losses almost do not decrease.
On the positive side, we do observe better test performance in some cases with soft derandomization included (see, e.g., the curves for $\tau = 0.1$).
\section{Discussion}

In this work, we study training-test (mis)alignment in unsupervised combinatorial optimization (UCO).
We identify a methodological misalignment issue in existing UCO methods, provide empirical evidence for the issue, and propose a preliminary idea for it.
Our idea of including soft derandomization into training appears to be a promising direction for further exploration.
Our analysis suggests that the future development of UCO may need to achieve better training-test alignment while maintaining stable training.

Beyond UCO, other machine learning methods for combinatorial optimization have also raised discussions on test-time post-processing~\citep{xia2024position}.
We believe researchers should be more careful about test-time post-processing in general, especially that they should be aware of the potential danger that too powerful test-time post-processing (compared to the training process) might make the training less relevant.
Especially, before addressing generalization regarding data distribution shift~\citep{luo2023heavydecoder}, researchers may need to first address the training-test methodological misalignment to ensure meaningful training.


\clearpage
\newpage

\normalem
\bibliographystyle{plainnat}
\bibliography{ref}

\end{document}